# Predicting e-commerce customer conversion from minimal temporal patterns on symbolized clickstream trajectories

Jacopo Tagliabue[1†], Lucas Lacasa[2], Ciro Greco[1], Mattia Pavoni[1] and Andrea Polonioli[1]

[1] Tooso Labs, San Francisco CA, USA
[2] Queen Mary University, London, United Kingdom
† Corresponding author.
`jacopo.tagliabue@tooso.ai`

**Abstract.** Knowing if a user is a "buyer" vs "window shopper" solely based on clickstream data is of crucial importance for e-commerce platforms seeking to implement real-time accurate NBA ("next best action") policies. However, due to the low frequency of conversion events and the noisiness of browsing data, classifying user sessions is very challenging. In this paper, we address the clickstream classification problem in the eCommerce industry and present three major contributions to the burgeoning field of A.I.-for-retail: first, we collected, normalized and prepared a novel dataset of live shopping sessions from a major European e-commerce website; second, we use the dataset to test in a controlled environment strong baselines and SOTA models from the literature; finally, we propose a new discriminative neural model that outperforms neural architectures recently proposed by [1] at Rakuten labs.

## 1 Introduction

The extraordinary growth of online distribution channels [2] has had a significant impact on the retail industry [3] [4]. However, a problem for digital retailers is that the vast majority of sessions are from users with weak buying intention ("window shoppers"). Being able to turn window shoppers into converting customers has thus become a key priority for clicks and mortar stores and solely digital players [5]. In turn, next-best-action marketing and personalization have recently become increasingly popular in an effort to increase conversion rates [6].

In *this* paper, we present the ongoing research that Tooso Labs is conducting on real-time intent detection, by marrying Artificial Intelligence and deep domain knowledge over proprietary eCommerce data. The paper is organized as follows: in Section 2 we define the intent detection problem; in Section 3 we detail our methodology and describe all the models in our study. In Section 4 we present results and a preliminary analysis before concluding, in Section 5, with some final remarks.

## 2   Problem Statement and Dataset

**2.1 Problem Statement**

The clickstream challenge is to predict if users on a website are likely/unlikely to buy *within* the session based solely on behavioral evidence (e.g. page view, search activity, etc.); as such it is usually framed as a *classification* problem, where the goal is to classify a session as BUY ("buy-session") or NOBUY ("no-buy-session"). To simplify the exploration of new methods and align with other literature baselines, we start with a version of the problem where the session is entirely available to the classifier, instead of data points being streamed in one at a time (mimicking real-time data streaming on browsing activities).

Formally, a session $s$ is a series of browsing events $e_1$, $e_2$, ...$e_n$ with timestamps $t_1$, $t_2$, ...$t_n$ by a user $u$, where the gap between any two times $t$ is at most 30 minutes [7]; events belong to different *categories*, such that each $e \in \{C \mid$ "view", "click", "detail", "add-to-cart", "remove-from-cart", "buy"$\}$. A session $s$ is classified as BUY ("buy-session") if and only if there is at least one event $e$ in $s$ such that $C_e =$ "buy", NOBUY ("no-buy-session") otherwise; to avoid trivializing the prediction problem, we cut sessions with "buy" events at the timestamp before the event.

**2.2 The Tooso Retail Clickstream Dataset**

Our "Tooso retail clickstream dataset" (TRCD) contains data from real user sessions on an e-commerce website from a major (>1B year turnover) retail group in Europe. All events in the dataset are sampled from the period from 06/29/2018 to 07/18/2018.

**2.3 The symbolized clickstream dataset**

We "symbolize" user sessions, so that, for each $e$ in $s$, the only information we retrieve is the event type. This simplifies the implementation of new algorithms, allows to readily make comparisons with SOTA models in the literature and make the findings imme-diately applicable to a wide range of use cases (in which detailed meta-data about events may be missing). After cleaning the raw datasets (excluding sessions shorter than 10 events and longer than 200 to avoid suspect sessions into the analysis), the final corpus consists of 7,176 BUY sessions and 123,396 NOBUY sessions.

## 3   Methodology

In what follows, we describe all the models and related implementation details.

**3.1 Literature comparisons**

*3.1.1 Markov chains*

We reproduce the methodology of the influential [8], which borrows from the long-standing idea that browsing activities are properly modelled as Markov processes [9] [10]. In particular, two separate Markov chains are trained for BUY/NOBUY sequences. At prediction time, for any session $s$, we predict the class associated with the highest probability, i.e. $s \in$ BUY iff $P(BUY|s) > P(NOBUY|s)$, $s \in$ NOBUY otherwise. We run several experiments to pick the best degree for the final chains and found that chains of order 5 provide the most reliable classification accuracy (in line with [1]).

*3.1.2 LSTM language model*

A recent paper [1] reported improvements over the MC approach in [8] using LSTMs [1]. While the authors frame the problem as a three-fold classification (*purchase*, *abandon* or *browsing-only*), they used the same idea of "mixture models" as in [8] just replacing MCs with probabilities from a neural network model (token probabilities are read off intermediate softmax layers in each LSTM model). We built two LSTMs (BUY, NOBUY) with as many input units as there are input classes and the same number of output units. We used Cross Entropy as our loss function and trained the network using Adam. In line with [1], we considered architectures with 1 hidden layer and 4 different values for the number of hidden units (10, 20, 40, 80); we also explored different values for the learning rate (0.01, 0.001) and for the batch size (10, 20, 50). We trained each model with early stopping on the accuracy on the validation set, with a patience of 10 and a maximum number of epochs of 50. At prediction time, each sequence is passed through both LSTMs and the probability of every state in the sequence is retrieved from the softmax layer. Classification happens in the same way as in the MC model.

**3.2 Novel contributions**

*3.2.1 Seq2Label*
We implemented a discriminative classifier as an alternative way to conceptualize the clickstream problem. This architecture consisted of one LSTM layer of dimensionality (input units x hidden units) and one fully connected layer with dimensionality (hidden units x 1), whose output was transformed using the sigmoid activation function before computing the loss. Two pooling strategies were explored, changing the information that is used to classify sequences: taking the output of the LSTM at the last time step (ignoring padding indices) and taking the average LSTM output over the entire sequence (again ignoring padding indices). The pooled output of the LSTM was then passed through the fully connected layer and transformed using the sigmoid activation function, which was then taken as the prediction given the sequence. Binary Cross Entropy Loss is used to quantify the error and back-propagate it (Adam was again the chosen optimizer). We tested the same hyper-parameter settings as in the LSTM language model and trained each model with early stopping, considering accuracy on the validation set as the target variable with a patience of 10.

*3.2.2 Visibility Graphs*

Leveraging the symbolic nature of the clickstream dataset, we explore a completely different prediction method by feeding *k*-grams to visibility graphs [11] [12]. The key intuition of this method is that you can induce a graph from a timeseries by linking events (as "nodes") that can "see" each other in the series, as shown in the figure below (see [12] for a formal introduction):

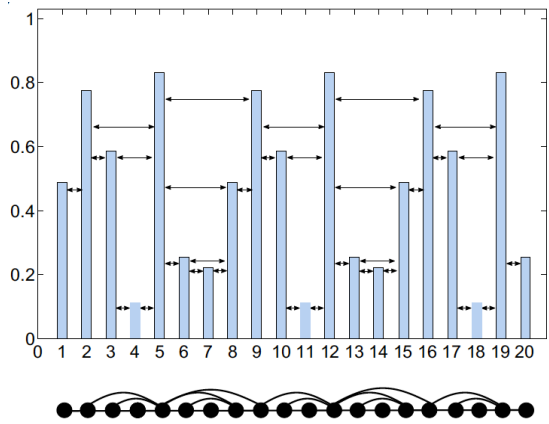

**Fig. 1.** Sample time series of 20 data and its associated horizontal visibility graph.

After transforming each symbolized session into the corresponding visibility graph, graph patterns are run through PCA to avoid overfitting and the resulting features are fed into a standard SVM classifier.

## 4  Results and Analysis

After deciding on the best parameter choices for our models, we tested them on the test split of the corpus (since LSTMs depend on random initialization, we trained 10 different instances of the same model). In the following table we report average accuracy scores on the test set and provide the standard deviation over 10 runs in parentheses when necessary.

**Table 1.** Average accuracy scores

| Model | Accuracy |
|:---:|:---:|
| Markov Chain | 0.882 |
| LSTM - Language Model | 0.909 (± 0.004) |
| Visibility Graphs | 0.868 (± 0.48) |
| LSTM - S2L ('avg' pooling) | 0.927 (± 0.003) |
| **LSTM - S2L ('last')** | **0.932 (± 0.002)** |

# 5 Conclusion

We presented preliminary but encouraging results in the clickstream prediction challenge for online retail. Using our novel dataset of live shopping sessions from a major European e-commerce website, we have proposed i) a new discriminative neural model that outperforms SOTA architectures proposed by [1]; ii) a physics-based approach, through visibility graphs, that can be thought as a very strong baseline for timeseries problems, being formally well understood, easy to implement and fast and cheap to compute even on large datasets. In the spirit of reproducibility, the authors plan to re-lease the full dataset and benchmarking code under a research-friendly license.